\documentclass[letterpaper,twocolumn,10pt]{article}
\usepackage{usenix2019_v3}
\usepackage{comment}

\usepackage{tikz}
\usepackage{amsmath}

\usepackage[numbers]{natbib}
\newlength{\bibitemsep}
\newlength{\bibparskip}
\let\oldthebibliography\thebibliography
\renewcommand\thebibliography[1]{%
  \oldthebibliography{#1}%
  \setlength{\parskip}{\bibitemsep}%
  \setlength{\itemsep}{\bibparskip}%
}

\begin{document}

\date{}

\title{\Large \bf Machine Learning Pipelines: Provenance, Reproducibility and FAIR Data Principles}
\vspace{-0.5cm}
\author{
Sheeba Samuel\href{https://orcid.org/0000-0002-7981-8504}{\includegraphics[scale=0.06]{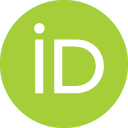}} 
\qquad
Frank L{\"o}ffler\href{https://orcid.org/0000-0001-6643-6323}{\includegraphics[scale=0.06]{orcid}} 
\qquad
Birgitta K\"onig-Ries\href{https://orcid.org/0000-0002-2382-9722}{\includegraphics[scale=0.06]{orcid}}\\
Heinz-Nixdorf Chair for Distributed Information Systems\\
Michael Stifel Center Jena\\
Friedrich Schiller University Jena\\
\rm{\{sheeba.samuel, frank.loeffler, birgitta.koenig-ries}\}@uni-jena.de
}
\maketitle

\begin{abstract}
Machine learning (ML) is an increasingly important scientific tool supporting decision making and knowledge generation in numerous fields. With this, it also becomes more and more important that the results of ML experiments are reproducible. Unfortunately, that often is not the case. Rather, ML, similar to many other disciplines, faces a reproducibility crisis.
In this paper, we describe our goals and initial steps in supporting the end-to-end reproducibility of ML pipelines.
We investigate which factors beyond the availability of source code and datasets influence reproducibility of ML experiments.
We propose ways to apply FAIR data practices to ML workflows.
We present our preliminary results on the role of our tool, ProvBook, in capturing and comparing provenance of ML experiments and their reproducibility using Jupyter Notebooks.
\end{abstract}
\section{Introduction}
Over the last few years, advances in artificial intelligence and machine learning (ML) have led to their use in numerous applications. With more and more decision making and knowledge generation being based on ML, it becomes increasingly important, that ML experiments are reproducible.
Only reproducible results are trustworthy and a suitable basis for future work.
Unfortunately, similar to other disciplines, ML faces a ``reproducibility crisis''~\cite{ Hutson725, samuel_thesis}.
In this paper, 
we investigate which factors contribute to this crisis and propose first solutions to address some of them.
We have conducted an initial study among domain experts for a better understanding of the requirements for reproducibility of ML experiments.
Based on the results from the study and the current scenario in the field of ML, we propose the application of FAIR data practices \cite{wilkinson2016fair} in end-to-end ML pipelines.
We use the ontologies to achieve interoperability of scientific experiments.
We demonstrate the use of ProvBook to capture and compare the provenance of executions of ML pipelines through Jupyter Notebooks.
Tracking the provenance of the ML workflow is needed for other scientists to understand how the results are derived.
Along with the provenance, the version for each provenance item needs to be maintained for the end-to-end reproducibility of an ML pipeline.
\vspace{-0.5cm}
\section{The situation: Characteristics of Machine Learning Experiments and their Reproducibility}
\vspace{-0.3cm}
An ML pipeline consists of a series of ordered steps used to automate the ML workflow.
Even though the general workflow is the same for most ML experiments, there are many activities, tweaks, and parameters that are involved which require proper documentation.
We conducted an internal study among 15 domain experts to understand what is needed to achieve reproducibility of ML experiments.
We asked questions regarding the reproducibility of results, the challenges in reproducing published results and the factors required for describing experiments for reproducibility in the field of ML.
We present here some relevant challenges and problems faced by scientists in reproducing published results of others:
(1) Unavailability, incomplete, outdated or missing parts of source code
(2) Unavailability of datasets used for training and evaluation
(3) Unavailability of a reference implementation
(4) Missing or insufficient description of hyperparameters that need to be set or tuned to obtain the exact results
(5) Missing information on the selection of the training, test and evaluation data
(6) Missing information of the required packages and their version
(7) Tweaks performed in the code not mentioned in the paper
(8) Missing information in methods and the techniques used, e.g., batch norm or regularization techniques
(9) Lack of documentation of preprocessing steps including data preparation and cleaning
(10) Difficulty in reproducing training of large neural networks due to hardware requirements. 
All the participants mentioned that if ML experiments are properly described with all the entities of the experiments and their relationships between each other, it will benefit them not only in the reproducibility of results but also for comparison to other competing methods (baseline).
The results of this survey are available online\footnote{\url{https://github.com/Sheeba-Samuel/MLSurvey}}.
\section{Towards a solution: Applying FAIR data principles to ML}
The FAIR data principles not only apply to research data but also to the tools, algorithms, and workflows that lead to the results.
This aids to enhance transparency, reproducibility, and reuse of the research pipeline.
In the context of the current reproducibility crisis in ML, there is a definite need to explore how the FAIR data principles and practices can be applied in this field.
In this paper, we focus on two of the four principles, namely Interoperability and Reusability which we equate with reproducibility.
To implement FAIR principles regarding interoperability, it is important that there is a common terminology to describe, find and share the research process and datasets.
Describing ML workflows using ontologies could, therefore, help to query and answer competency questions like:
(1) Which hyperparameters were used in one particular run of the model?
(2) Which libraries and their versions are used in validating the model?
(3) What is the execution environment of the ML pipeline?
(4) How many training runs were performed in the ML pipeline?
(5) What is the allocation of samples for training, testing and validating the model?
(6) What are the defined error bars?
(7) Which are the measures used for evaluating the model?
(8) Which are the predictions made by the model?

In previous work, we have developed the REPRODUCE-ME ontology which is extended from PROV-O and P-Plan~\cite{samuel_thesis}. REPRODUCE-ME introduces the notions of Data, Agent, Activity, Plan, Step, Setting, Instrument, and Materials, and thus models the general elements of scientific experiments required for their reproducibility.
Work is in progress to extend this ontology to include ML concepts which scientists consider important according to our survey. We also aim to be compliant with existing ontologies like ML-Schema \cite{MLSchema} and MEX vocabulary \cite{MEXVocabulary}.
With REPRODUCE-ME, the ML pipeline developed through Jupyter Notebooks can be described in an interoperable manner.
\section{Achieving Reproducibility using ProvBook}
Building an ML pipeline requires constant tweaks in the algorithms and models and parameter tuning.
Training of the ML model is conducted through trial and error.
The role of randomness in ML experiments is big and its use is common in steps like data collection, algorithm, sampling, etc.
Several runs of the model with the same data can generate different results.
Thus, repeating and reproducing results and reusing pipelines is difficult.

The use of Jupyter Notebooks is rapidly increasing as they allow scientists to perform many computational activities including statistical modeling, machine learning, etc.
They support computational reproducibility by allowing users to share code along with documentation and results.
However, the surveys~\cite{Pimentel2019} on Jupyter Notebooks point out the need of provenance information of the execution of these notebooks.
To overcome this problem, we developed ProvBook \cite{samuel_thesis}.
With ProvBook, users can capture, store, describe and compare the provenance of different executions of Jupyter notebooks.
This allows users to compare the results from the original author with their own results from different executions.
ProvBook provides the difference in the result of the ML pipeline from the original author of the Jupyter notebook in GitHub with the result from our execution using ProvBook.
Even though the code and data remain the same in both the executions, there is a subtle difference in the result.
In ML experiments, users need to figure out the reason behind different results due to modification in data or models or because of a random sample.
Therefore, it is important to describe the data being used, the code and parameters of the model, the execution environment to know how the results have been derived.
ProvBook helps in achieving this reproducibility level by providing the provenance of each run of the model along with the execution environment.
\vspace{-0.5cm}
\section*{Acknowledgments}
The authors thank the Carl Zeiss Foundation for the financial support of the project ``A Virtual Werkstatt for Digitization in the Sciences (K3)'' within the scope of the program-line ``Breakthroughs: Exploring Intelligent Systems'' for ``Digitization -- explore the basics, use applications''.
\vspace{-0.5cm}
\bibliographystyle{unsrt}
\bibliography{poster}

\end{document}